\documentclass{article}

\usepackage{microtype}
\usepackage{graphicx}
\usepackage{subfigure}
\usepackage{booktabs} 

\usepackage{hyperref}

\usepackage[accepted]{icml2021}

\usepackage{url}
\usepackage[utf8]{inputenc} 
\usepackage[T1]{fontenc}    
\usepackage{booktabs}       
\usepackage{amsfonts}       
\usepackage{nicefrac}       
\usepackage{microtype}      
\usepackage{amsthm}
\usepackage{epstopdf}
\usepackage{bm,xspace}
\usepackage{verbatim}
\usepackage{multirow}
\usepackage{wrapfig}
\usepackage{balance}
\usepackage{calc}
\usepackage{framed}
\usepackage{eqparbox}
\usepackage{xspace}
\usepackage{xr}
\usepackage{xcolor}
\usepackage{empheq}
\usepackage[most]{tcolorbox}

\usepackage{siunitx} 

\newcommand{\llama}{LLaMA}
\newcommand{\neft}{\texttt{NEFT}}
\newcommand{\neftune}{\texttt{NEFTune}}
\newcommand{\symnoise}{\texttt{SymNoise}}

\begin{document}

\twocolumn[
\icmltitle{\symnoise{}:~Advancing Language Model Fine-tuning with Symmetric Noise}




\icmlsetsymbol{equal}{*}

\begin{icmlauthorlist}
\icmlauthor{Abhay Kumar Yadav}{goo}
\icmlauthor{Arjun Singh}{to}\\
\icmlauthor{University of Maryland, College Park, MD, USA}{goo}
\icmlauthor{Savitribai Phule Pune University, Pune, India}{to}
\end{icmlauthorlist}

\icmlaffiliation{to}{Savitribai Phule Pune University, Pune, India}
\icmlaffiliation{goo}{University of Maryland, College Park, MD, USA}

\icmlcorrespondingauthor{Arjun Singh}{arjun.singh071954@gmail.com}

\icmlkeywords{Machine Learning, ICML}

\vskip 0.3in
]





\begin{abstract}
In this paper, we introduce a novel fine-tuning technique for language models, which involves incorporating symmetric noise into the embedding process. This method aims to enhance the model's function by more stringently regulating its local curvature, demonstrating superior performance over the current method, \neftune{}. When fine-tuning the  \llama{}-2-7B model using Alpaca, standard techniques yield a $29.79$\% score on AlpacaEval. However, our approach, \symnoise{}, increases this score significantly to $69.04$\%, using symmetric noisy embeddings. This is a $6.7$\% improvement over the state-of-the-art method, \neftune{}~($64.69$\%). Furthermore, when tested on various models and stronger baseline instruction datasets, such as Evol-Instruct, ShareGPT, OpenPlatypus, \symnoise{} consistently outperforms \neftune{}.  The current literature, including \neftune{}, has underscored the importance of more in-depth research into the application of noise-based strategies in the fine-tuning of language models. Our approach, \symnoise{}, is another significant step towards this direction, showing notable improvement over the existing state-of-the-art method.
\end{abstract}

\section{Introduction}
For Large Language Models (LLMs) to be effective, their proficiency in executing specific instructions is crucial~\cite{wang2022self-selfinstruct, ouyang2022training-instructGPT, brown2020language-gpt3, chung2022scaling-FLAN}. These models typically begin with training on a vast array of unfiltered web data, after which they undergo a more focused fine-tuning stage using a smaller, selectively chosen collection of instructional data. The fine-tuning stage, centered on instructions, is fundamental in unlocking and controlling the full capabilities of LLMs. The practical value of these models is predominantly dependent on how efficiently we can leverage these concise instructional data sets for optimal performance.

Recent advancements in computer vision have thoroughly investigated methods involving noise injection~\cite{nukrai2022text-cap, zang2021noise-pointcloud, akbiyik2020data-noiseimage}, and these approaches are now gaining traction in the fine-tuning of Large Language Models (LLMs) due to their simplicity and effectiveness in enhancing performance. A notable example is the \neftune{} method~\cite{jain2023neftune}, which incorporates uniform random noise. While \neftune{} significantly surpasses the performance of standard fine-tuning methods, its underlying mechanics remain somewhat enigmatic. In this paper, we introduce an alternative noise injection technique that not only offers an intuitive understanding but also demonstrates superior performance, surpassing both \neftune{} and other existing methodologies by a considerable margin.

To this end, we introduce a novel noise type that not only exhibits superior performance but is also founded on a more profound understanding. Our objective is to regularize the curvature of the function learned during training. Curvature regularization has been used in domains such as computer vision~\cite{moosavi2019robustness, lee2023explicit}, graph embedding~\cite{pei2020curvature}, and  deep neural networks~\cite{huh2020curvature}. Specifically, we aim to ensure that the function's response changes gradually when the input is modified slightly by noise. In more technical terms, our goal is to have the gradient approach zero in the immediate vicinity of an input altered by a minimal amount. This represents a more stringent condition than merely requiring small values for the Hessian or gradient. However, considering computational efficiency, we opt to avoid the direct computation of gradients or Hessians. Instead, we employ this stringent condition, which, as our experiments on real-world benchmark datasets demonstrate, is effective in practical scenarios.

In this paper, we unveil Symmetric Noise Fine Tuning (\symnoise{}), a new technique that leverages symmetric Bernoulli distribution-based noise applied to the embedding vectors of training data during the finetuning stage. Each noise component is generated with an equal probability of $\frac{1}{2}$ for the values $-1$ and $1$. This method significantly enhances instruction finetuning outcomes, often with remarkable gains, while avoiding additional computational or data resources. While maintaining simplicity, \symnoise{} has a profound impact on downstream conversational output quality. We show that when a large langudage model like \llama{}-2-7B~\cite{touvron2023llama-2} is finetuned using \symnoise{}, its performance on \texttt{AlpacaEval}~\cite{dubois2023alpacafarm-rlhf} rises from $29.79$\% to $69.04$\% – a substantial increase of about $39.25$ percentage points.

Importantly, when compared to the existing \neftune{} method~(which uses random uniform noise), \symnoise{} demonstrates a superior performance edge, outperforming \neftune{} by approximately $6.7$\%. Thus, \symnoise{} not only represents a valuable advancement over traditional finetuning methods but also establishes a new benchmark in efficiency and effectiveness for LLM finetuning.

\begin{table*}[t]
\centering
\caption{\texttt{AlpacaEval} Win Rate against Text-Davinci-003 when applied with \llama{}-2, trained across diverse datasets. Utilizing GPT for evaluation, \symnoise{} shows an overall improvement throughout all datasets, outperforming \neftune{} on all datasets~(as reported in~\citet{jain2023neftune}).}\label{tab:LLaMA-2_GPT}
\begin{tabular}{|l|c|c|c|c|c|}
\toprule
            & Alpaca & Evol-Instruct & ShareGPT & OpenPlatypus & Average \\ \midrule
\llama{}-2 7B  & 29.79  & 70.34    & 68.74              & 62.00        & 57.71   \\ 
+\neft{}  & 64.69 & 79.60    & 76.28              & 70.61        & 72.80  \\ 
+\symnoise{}  & \textbf{69.04}  & \textbf{81.38}    & \textbf{78.67}              & \textbf{72.23}        & \textbf{75.33}  \\ \bottomrule
\end{tabular}
\end{table*}

\begin{table*}[ht]
\centering
\caption{\texttt{AlpacaEval} Win Rate with and without \neftune{}, \symnoise{} on \llama{}-2, \llama{}-1, and OPT on datasets including Alpaca, Evol-Instruct, ShareGPT, and OpenPlatypus. Notably, there's an enhancement in performance using \symnoise{} over \neft{}~across these datasets and models when evaluated using ChatGPT. For completeness, results for base models and \neft{} are reported from~\citet{jain2023neftune}}\label{table-fig2}
\begin{tabular}{|l|S|S|S|S|}
\hline
Method/Dataset & {Alpaca} & {Evol-Instruct} & {OpenPlatypus} & {ShareGPT} \\ \hline
\hline
OPT-6.7B               & 41.4 & 52.2 & 45.7 & 53.4 \\ 
OPT-6.7B +\neft{}         & 48.7 & 55.5 & 45.8 & 54.3 \\ 
OPT-6.7B +\symnoise{}     & 50.8 & 57.6 & 46.9 & 55.6 \\ \hline
\hline
LLaMA-1-7B             & 48.5 & 62.3 & 51.2 & 62.9 \\ 
LLaMA-1-7B +\neft{}       & 61.7 & 67.5 & 56.9 & 63.6 \\ 
LLaMA-1-7B +\symnoise{}   & 64.0  & 69.8 & 58.5  & 65.4 \\ \hline
\hline
LLaMA-2-7B             & 48.3 & 62.5 & 57.2 & 63.5 \\ 
LLaMA-2-7B +\neft{}       & 62.5 & 67.6 & 61.7 & 64.2 \\ 
LLaMA-2-7B +\symnoise{}   & 64.9 & 69.6 & 62.1 & 66.1 \\ 
\hline
\end{tabular}
\end{table*}

\section{Background and Related Work}
In the evolving landscape of instruction finetuning for Large Language Models (LLMs), initial efforts like FLAN and T0 marked the beginning of cross-task generalization~\cite{sanh2021multitask-T0, wei2021finetuned-FLAN}. These models, involving encoder-decoder language architectures, underwent finetuning across a diverse spectrum of thousands NLP tasks. This progression, detailed in studies by~\citet{chung2022scaling-FLAN} and \citet{xu2022zeroprompt} demonstrated the adaptability of LLMs to a variety of standard NLP tasks.

Following this trajectory, OpenAI's InstructGPT~\cite{ouyang2022training-instructGPT} emerged as a pioneering model adept at handling open-ended questions with remarkable efficiency. This model, an iteration of GPT-3~\cite{brown2020language-gpt3}, incorporated reinforcement learning from human feedback (RLHF), leading to the development of advanced models like ChatGPT~\cite{OpenAI2022ChatGPT}. ChatGPT, in particular, gained widespread attention for generating more coherent and extended texts compared to InstructGPT.

Building on these developments,~\citet{wang2022self-selfinstruct} introduced the Self-Instruct approach, utilizing InstructGPT to generate instructional pairs for further finetuning of foundational models like LLaMA into specialized variants such as Alpaca~\cite{taori2023stanford-alpaca}. Concurrently, the trend towards distilled models, as discussed by~\citet{taori2023stanford-alpaca} and~\citet{xu2023wizardlm}, led to the creation of diverse datasets. These datasets, including works by~\citet{xu2023wizardlm} and~\citet{lee2023platypus}, focused on refining specific model capabilities like STEM question answering and logical reasoning. Another notable advancement was AlpaGasus by~\citet{chen2023alpagasus}, which employed a quality-filtering mechanism based on GPT-4 evaluations to enhance model performance. In a different methodology, ShareGPT, as described by \citet{Chiang2023Vicuna}, was developed through the crowd sourcing of real user interactions sourced from ChatGPT.

In the context of incorporating noise into model training, the pioneering work by~\citet{zhu2019freelb} with the FreeLB method demonstrated the effectiveness of adversarial perturbations in boosting MLM model performance. This method involved introducing calculated Gaussian perturbations into the embeddings and optimizing them to maximally impact model performance. Similar strategies were later applied in various domains, such as image captioning~\cite{nukrai2022text-cap}, point cloud processing~\cite{zang2021noise-pointcloud}, graphs~\cite{kong2022robust-adversarialnoise}, and privacy mechanisms~\cite{dwork2014algorithmic-diffprivacy}. Curvature regularization has been used in domains such as computer vision~\cite{moosavi2019robustness, lee2023explicit}, graph embedding~\cite{pei2020curvature}, and  deep neural networks~\cite{huh2020curvature}. Noise based on the Bernoulli distribution, as opposed to Gaussian or Uniform noise, has been utilized, as mentioned by \citet{spall1998implementation}. In this approach, each outcome, either $-1$ or $1$, is assigned an equal probability of $\frac{1}{2}$.

\section{Proposed Method: \symnoise{}}
In the ideal scenario, our goal is to implement curvature regularization, a technique prevalent in fields such as computer vision~\cite{moosavi2019robustness, lee2023explicit}, graph embedding~\cite{pei2020curvature}, and deep neural networks~\cite{huh2020curvature}. However, due to the high computational cost associated with these methods, we aim to explore an alternative approach that adheres to a more stringent condition. This approach has demonstrated superior performance in practice, surpassing current state-of-the-art methodologies. Specifically, we seek to design a function with a gradient~($\nabla f$) having value as $0$ in the vicinity of the input, i.e., \text{for} $x, \epsilon \in \mathbb{R}^{d}$,\\

$\nabla f = \frac{\left|{\frac{f{\left((x - \epsilon ) \right)} - f{\left(x + \epsilon \right)}}{\epsilon}}\right|}{2} \leq \delta $, when $\delta=0$, we have $f{\left(x + \epsilon  \right)} = f{\left(x - \epsilon  \right)}.$\\
\\
Our approach is intuitively designed to introduce a noise vector that is the exact inverse of Bernoulli vectors, thereby compelling the network to learn an identical function for both points of symmetric perturbation around the input vector. In this formulation, the noise turns out to be based on a Bernoulli distribution, diverging from the more commonly used Gaussian or Uniform noise types. This method involves the application of Bernoulli noise with an equal probability of $\frac{1}{2}$ for each possible outcome, specifically $-1$ and $1$, as implemented in the methodology described by~\citet{spall1998implementation}. This choice is driven by the aim to achieve a more balanced and predictable influence on the learning process of the network.

Following the approach outlined by \citet{jain2023neftune}, we conduct the training of instruction-based models using datasets that consist of pairs of instructions and their corresponding responses. Our methodology diverges from \neftune{}, which typically begins by sampling an instruction, converting its tokens into embedding vectors, and then introducing uniform random noise to these embeddings. In our approach, while we retain the noise scaling factor $\epsilon = \alpha / \sqrt{Ld}$, as utilized in \neftune{} and similar works by Zhu et al. and Kong et al., where $L$ denotes the sequence length, $d$ the embedding dimension, and $\alpha$ a tunable parameter, we differ in the noise application process and use symmetric opposing Bernoulli noise as explained above. The specifics of our proposed method~\symnoise{} are elucidated in Algorithm~\ref{alg:symnoise}, with \neftune{} is presented in Algorithm~\ref{alg:NEFTune} for comparative analysis.

\begin{algorithm*}[ht]
   \caption{\neftune{}: \textbf{N}oisy \textbf{E}mbedding Instruction \textbf{F}ine\textbf{tun}ing~(Taken from the paper~\cite{jain2023neftune})} \label{alg:NEFTune}
\vspace{0.2cm}
\begin{minipage}{\textwidth}
\begin{algorithmic}
\STATE \textbf{Input:} $\mathcal{D}=\{x_i,y_i\}_1^N$ tokenized dataset, embedding layer $\text{emb}(\cdot)$, rest of model $f_{/\text{emb}}(\cdot)$, \\ model parameters $\theta$, $\text{loss}(\cdot)$, optimizer $\text{opt}(\cdot)$
\STATE Hyperparameter: base noise scale $\alpha \in \mathbb{R^+}$
\STATE Initialize $\theta$ from a pretrained model.
\vspace{0.2cm}
\REPEAT
\STATE $(X_i,Y_i) \sim \mathcal{D}$
\COMMENT{sample a minibatch of data and labels}
\STATE $X_{\text{emb}} \gets \text{emb}(X_i), \mathbb{R}^{B\times L \times d}$
\COMMENT{batch size $B$, seq. length $L$, embedding dimension $d$}
\STATE $\epsilon \sim \text{Uniform}(-1,1), \mathbb{R}^{B\times L \times d}$ \COMMENT{sample a noise vector}
\STATE $X_{\text{emb}}' \gets X_{\text{emb}} + (\frac{\alpha}{\sqrt{Ld}}) \epsilon $ \COMMENT{add scaled noise to embeds \footnote{If sequence lengths in a batch are not equivalent, then $L$ is a vector $\in \mathbb{Z}_{>0}^{B}$ and the scaling factor $(\alpha/\sqrt{Ld})$ is computed independently for each sequence in batch.}}
\STATE $\hat{Y}_i \gets f_{/\text{emb}}(X_{\text{emb}}')$ \COMMENT{make prediction at noised embeddings}
\STATE $\theta \gets \text{opt}(\theta, \text{loss}(\hat{Y}_i,Y_i)) $ \COMMENT{train step, e.g., grad descent}
\UNTIL{Stopping criteria met/max iterations.}
\end{algorithmic}
\end{minipage}
\end{algorithm*}

\begin{algorithm*}[ht]
   \caption{\symnoise{}: \textbf{Sym}metric \textbf{Nois}y \textbf{E}mbedding Instruction Finetuning~(Proposed Method)} \label{alg:symnoise}
\vspace{0.2cm}
\begin{minipage}{\textwidth}
\begin{algorithmic}
\STATE \textbf{Input:} $\mathcal{D}=\{x_i,y_i\}_1^N$ tokenized dataset, embedding layer $\text{emb}(\cdot)$, rest of model $f_{/\text{emb}}(\cdot)$, \\ model parameters $\theta$, $\text{loss}(\cdot)$, optimizer $\text{opt}(\cdot)$
\STATE  Hyperparameter: base noise scale $\alpha \in \mathbb{R^+}$
\STATE Initialize $\theta$ from a pretrained model.
\vspace{0.2cm}
\REPEAT
\STATE $(X_i,Y_i) \sim \mathcal{D}$
\COMMENT{sample a minibatch of data and labels}
\STATE $X_{\text{emb}} \gets \text{emb}(X_i), \mathbb{R}^{B\times L \times d}$
\COMMENT{batch size $B$, seq. length $L$, embedding dimension $d$}
\STATE $\epsilon \sim \text{Bernoulli}\{-1,1\}, \mathbb{R}^{B\times L \times d}$ \COMMENT{sample a noise vector}
\STATE $X_{\text{emb}}' \gets X_{\text{emb}} + (\frac{\alpha}{\sqrt{Ld}}) \epsilon $ \COMMENT{add scaled noise to embeds \footnote{If sequence lengths in a batch are not equivalent, then $L$ is a vector $\in \mathbb{Z}_{>0}^{B}$ and the scaling factor $(\alpha/\sqrt{Ld})$ is computed independently for each sequence in batch.}}
\STATE $X_{\text{emb}}'' \gets X_{\text{emb}} - (\frac{\alpha}{\sqrt{Ld}}) \epsilon $ \COMMENT{subtract same symmetric noise from embeds}

\STATE $\hat{Y}_i \gets f_{/\text{emb}}(concat(X_{\text{emb}}', X_{\text{emb}}''))$ \COMMENT{make prediction at noised embeddings}
\STATE $\theta \gets \text{opt}(\theta, \text{loss}(\hat{Y}_i,Y_i)) $ \COMMENT{train step}
\UNTIL{Stopping criteria met/max iterations.}
\vspace{0.1cm}
\end{algorithmic}
\end{minipage}
\end{algorithm*}

\section{Experiments}

In this section, we perform numerous experiments across various models and benchmarks to demonstrate the efficacy of our proposed method~\symnoise{} and compare it with existing approaches including~\neft{}.

\subsection{Datasets}
In this section, we delve into finetuning datasets that have either gained widespread popularity or have recently achieved state-of-the-art results. Due to the memory limitations of our hardware setup, our focus is exclusively on single-turn datasets follwoing similar protcol as used in~\citet{jain2023neftune}. The chosen datasets are:

\begin{itemize}
  \item \textbf{Alpaca~\cite{taori2023stanford-alpaca}}: Developed using the Self-Instruct method by~\citet{wang2022self-selfinstruct} and the Text-Davinci-003~\citet{ouyang2022training-instructGPT} model (Ouyang et al., 2022), Alpaca leverages a small set of seed tasks to generate new instruction tuning tasks and filter out ineffective ones. This dataset has been instrumental in advancing instruction-based learning.
    \item \textbf{ShareGPT~\cite{Chiang2023Vicuna}}: Comprising 70k voluntarily shared ChatGPT conversations~\cite{ShareGPT2023}, ShareGPT provides a rich source of real-world interaction data. While originally multi-turn, we adapt it to a single-turn format using the Vicunav1.1 dataset version for consistency with our experimental setup.
  \item \textbf{Evol-Instruc~\cite{xu2023wizardlm}}: This dataset, comprising 70k single-turn instructions, is considered more complex than Alpaca. Originating from the Alpaca dataset, Evol-Instruct employs ChatGPT to refine and evolve the initial instructions, thus broadening the scope and complexity of the tasks.
  \item \textbf{Open-Platypus~\cite{lee2023platypus}}: Formed by combining $11$ open-source datasets, Open-Platypus is tailored to enhance LLM performance in STEM and logical reasoning domains. It includes approximately $25$k questions, with around $10$\% generated by LLMs and the rest by human experts. This dataset emphasizes the importance of variety and complexity in question formats.

\end{itemize}

In the fine-tuning phase, each model, with the exception of ShareGPT, utilizes the prompt from the Alpaca system. Conversely, ShareGPT is fine-tuned using the prompt from the Vicuna system. The specific hyperparameters employed are comprehensively detailed in Appendix A.1. Our approach to hyperparameter tuning, including the selection of values, aligns with the methodologies suggested by \citet{jain2023neftune}.

\subsection{Models}
Following~\citet{jain2023neftune} setup for experimentation, our experiments predominantly utilize Large Language Models (LLMs) with a parameter size of $7$ billion. Specifically, our focus is on models such as \llama{}-1~\cite{touvron2023llama-1}, \llama{}-2~\cite{touvron2023llama-2}, and OPT-6.7B~\cite{zhang2022opt}. These transformer-based models primarily differ in the amount of training data they've been exposed to, with OPT-6.7B, \llama{}-1, and \llama{}-2 being trained on $180$ billion, $1$ trillion, and $2$ trillion tokens, respectively. This variance in training data volume is expected to manifest in their performance across benchmarks like MMLU, where \llama{}-2 typically outperforms the others. Additionally, for models with 13B and 70B parameters, our focus is on training LLaMA-2.

\subsection{Evaluation Protocols}
Our experimental framework, adapted from the original \neftune{}~\cite{jain2023neftune} setup, primarily utilizes single-turn data for training. We assess the models' conversational skills using \texttt{AlpacaEval} and examine their performance on tasks from the OpenLLM Leaderboard. This was done to verify that the introduction of our \texttt{symnoise} augmentation does not negatively impact the models' performance on standard multiple-choice tasks. Notably, the results demonstrate that our augmented models consistently outperform the original \texttt{neftune} models, albeit by a modest margin.

\begin{itemize}
    \item \textbf{AlpacaEval}: Introduced by~\citet{dubois2023alpacafarm-rlhf}, \texttt{AlpacaEval} is crucial for appraising generative quality. It functions as an automated model-based evaluator, comparing Text-Davinci-003's generations with our model's over $805$ prompts, focusing on the \textit{Win Rate}. The Win Rate indicates how often our model is preferred over Text-Davinci-003, as judged by model evaluator~(GPT-4, ChatGPT etc). The dataset's $805$ test prompts, sourced from various platforms, ensure a comprehensive testing scope. \texttt{AlpacaEval's} high human agreement rate~\cite{dubois2023alpacafarm-rlhf}, validated by 20K annotations, highlights its usefulness and accuracy. We employ both GPT-4 and ChatGPT as evaluators, using ChatGPT initially due to GPT-4's API limitations and costs.

    \item \textbf{Hugging Face OpenLLM Leaderboard}: For leaderboard assessments, datasets like ARC~\cite{clark2018think-arc}, HellaSwag~\cite{zellers2019hellaswag-swag}, and MMLU~\cite{hendrycks2020measuring-mulu} are utilized. These verbalized multiclass classification datasets test the LLM's capability in factual questioning and reasoning. Our evaluations confirm that the \symnoise{} method does not diminish the models' proficiency in these domains.
\end{itemize}

\subsection{Results}
\subsubsection{Improvement in generated text quality}
Our results demonstrate an enhanced metric performance compared to~\neftune{} in terms of generated text quality. As evident from Table~\ref{tab:LLaMA-2_GPT}, there is a notable improvement across all datasets at the 7B size, with an average increase of $17.6\%$ (compared to~\neftune{}'s improvement of $15.1\%$). This indicates that the implementation of \symnoise{} significantly enhances conversational capabilities and answer quality. These findings are supported by evaluations using \texttt{AlpacaEval}, where \symnoise{} notably outperforms \neftune{}. Furthermore, as shown in Table \ref{table-fig2}, enhancements are also observed in older models like \llama{}-1 and OPT-6.7B, with \symnoise{} consistently surpassing \neftune{} in these models as well. An interesting observation is the comparatively smaller gain by \neftune{} in ShareGPT, as per ChatGPT's analysis, a trend not mirrored in GPT-4's evaluation. However, \symnoise{} consistently excels over \neftune{} for ShareGPT in evaluations by both GPT-4 and ChatGPT. In Table \ref{tab:LLaMA-2_GPT}, the Win Rate shows a significant increase from $29.79\%$ to $69.04\%$ for Alpaca, thereby outperforming the state-of-the-art method \neftune{} by $6.7\%$.

\subsubsection{Improvement in \textit{OpenLLM Leaderboard} tasks}
\begin{figure}[htbp]
    \centering
    \begin{subfigure}{}
        \includegraphics[width=\linewidth]{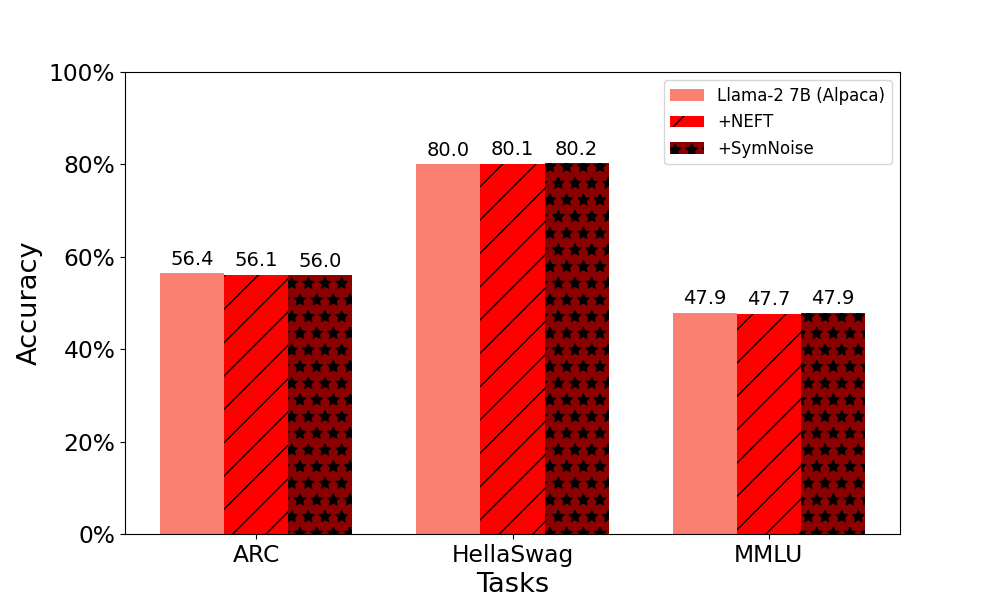}
    \end{subfigure}
    \\
    \begin{subfigure}{}
        \includegraphics[width=\linewidth]{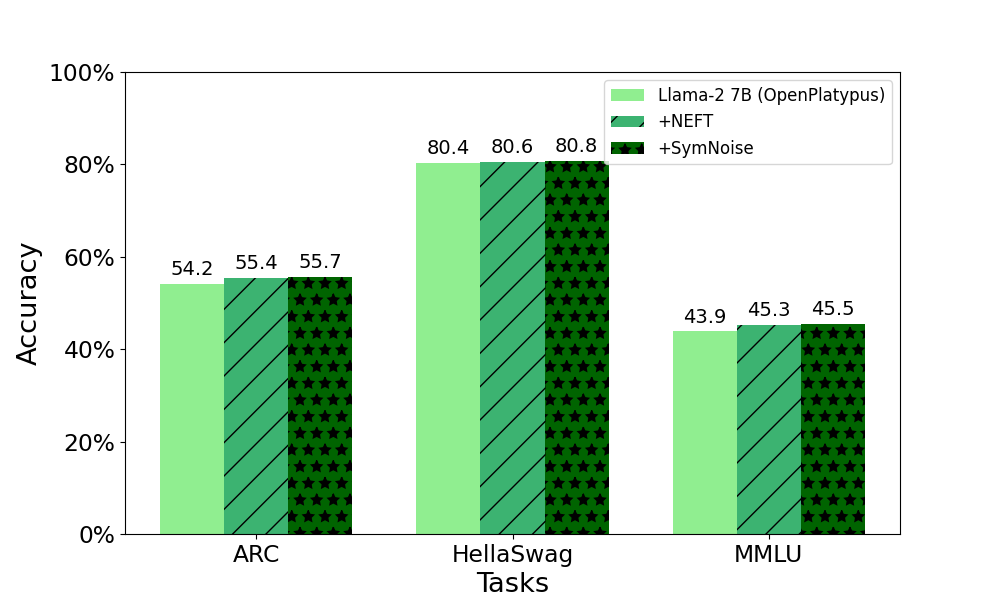}
    \end{subfigure}
    \\
    \begin{subfigure}{}
        \includegraphics[width=\linewidth]{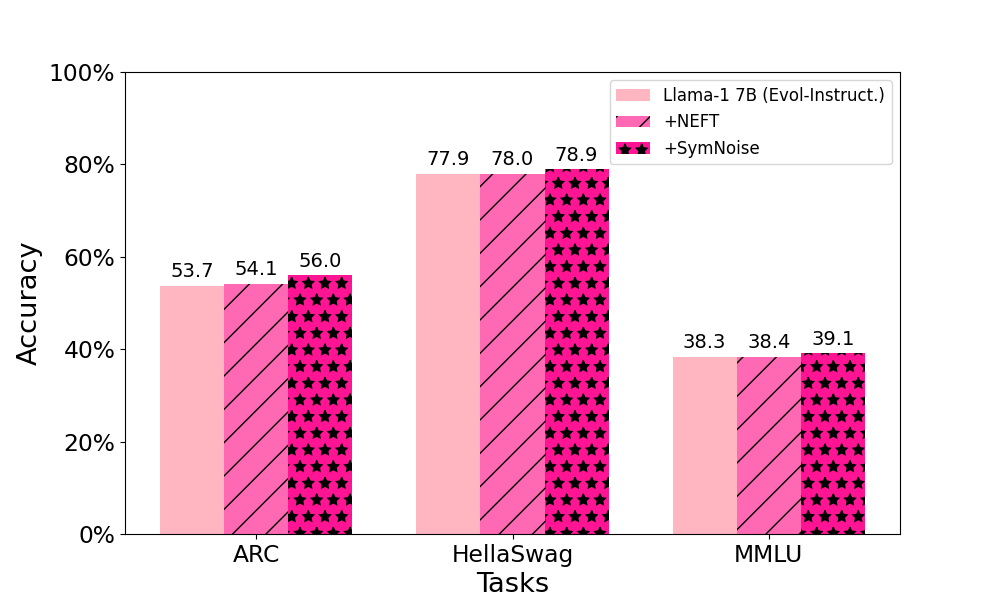}
    \end{subfigure}
    \caption{For OpenLLM Leaderboard tasks, the influence of \neftune{} and \symnoise{} is investigated on \llama{}-2, encompassing Alpaca, Evol-Instruct, and OpenPlatypus datasets, alongside \llama{}-1 trained on the Evol-Instruct dataset. Comparative observations reveal a uniformity in performance metrics across the diverse datasets and models, indicating negligible impact of \neftune{} but slightly better performance of \symnoise{} on the overall effectiveness. We follow the similar procedure as mentioned in ~\citet{jain2023neftune}, and report their results for completeness. In order to minimize computational expenses, we refrained from conducting thorough hyper-parameter optimization, which may have further improved the results.}
    \label{fig:fig3_openllm}
\end{figure}

In addressing the potential that \symnoise{} could enhance conversational abilities at the expense of traditional skills, we conducted evaluations using tasks from the OpenLLM Leaderboard. Employing the LM-Eval Harness framework \cite{gao2021framework-lm-eval-harness}, we assessed our model's performance on benchmarks such as MMLU~\cite{hendrycks2020measuring-mulu}, ARC~\cite{clark2018think-arc}, and HellaSwag~\cite{zellers2019hellaswag-swag}. These tests shed light on the model’s knowledge base, reasoning capabilities, and adherence to factual information. As illustrated in Figure \ref{fig:fig3_openllm}, the results indicate that \symnoise{} not only stabilizes scores but also actively preserves and, in some cases, enhances the model's capabilities. Notably, \symnoise{} consistently outperforms \neftune{} in terms of performance, highlighting its effectiveness in striking a balance between conversational proficiency and traditional computational skills.

\subsubsection{A Qualitative Example.}\label{sec:Qualitative_Example}
In this section, we present a specific example from \llama{}-2, trained on the Alpaca dataset, to compare the outcomes with and without the application of \symnoise{} and \neftune{}. To ensure objectivity, we used the same example as the one chosen in the \neftune{} paper \cite{jain2023neftune}, thus avoiding any bias in selection. Specifically, we selected the \emph{quantum computing prompt} from the Vicuna Eval subset within the \texttt{AlpacaEval} framework. This example demonstrates that the Alpaca model, in its basic form, provides a succinct response, briefly covering key aspects of quantum computing such as qubits, superposition, and entanglement, and their significance in complex computations. In contrast, when enhanced with \symnoise{}, the model offers a more detailed and articulate explanation. The augmented response delves deeper into the principles of superposition and quantum entanglement and also explores potential real-world applications. Interestingly, the response generated with \symnoise{} closely mirrors that of \neftune{}, yet it is more concise and uses fewer characters. This instance exemplifies the qualitative and quantitative improvements brought about by \symnoise{}, reflecting the broader enhancements observed across various fine-tuning datasets.

\begin{tcolorbox}[boxrule=0pt] 
  \textbf{\texttt{AlpacaEval} (Vicuna Subset) Prompt}: Can you explain the basics of quantum computing?\\
  \textbf{Alpaca-2-7B Response}: \\Quantum computing is based on the principles of quantum mechanics, which allow for phenomena such as superposition and entanglement between particles. This means that information can be processed in parallel instead of sequentially like traditional computers. Quantum bits (qubits) are used to store data instead of classical bits, allowing more complex computations to take place with fewer resources than would otherwise be necessary. \\
  \textbf{Alpaca-2-7B-\neft{} Response}:\\Quantum computing is based on the principles of quantum mechanics, which describe how particles such as electrons and photons behave at the subatomic level. Unlike traditional computers, which process information in binary digits (bits) that can be either 0 or 1, quantum bits (qubits) can exist in multiple states between 0 and 1 simultaneously. This phenomenon is known as superposition and allows for much more complex calculations to be performed than would otherwise be possible with classical computers. Quantum algorithms are designed to take advantage of this property by encoding data into qubit states so that it can be processed exponentially faster than conventional methods. Additionally, quantum entanglement--the ability for two particles to remain connected even over great distances--allows for large amounts of data to be sent across networks quickly and securely. The potential applications of quantum computing range from cryptography and artificial intelligence to drug discovery and material science. With continued research and development, quantum technology has the potential to revolutionize many industries. \\

    \textbf{Alpaca-2-7B-\symnoise{} Response}: \\Quantum computing, a leap from traditional computing, operates with quantum bits or qubits. Unlike classical bits that are either 0 or 1, qubits can exist in multiple states simultaneously, a phenomenon known as superposition. This allows quantum computers to process data in parallel, drastically differing from the sequential data processing of classical systems. Additionally, quantum entanglement, where qubits remain interconnected regardless of distance, enhances computational power and data security. These unique features enable quantum algorithms to solve complex problems much more efficiently than classical methods. Quantum computing's potential applications, ranging from advanced cryptography to breakthroughs in material science and drug discovery, illustrate its transformative capacity in various industries. \\
    
\end{tcolorbox}

\subsection{Analysis}

As highlighted in several studies, including \neftune{} \cite{jain2023neftune}, introducing noise into embeddings during training can reduce a model's tendency to overfit the idiosyncrasies of the instruction-tuning dataset, such as specific formatting, exact word choices, or text length. This approach shifts the model's focus from merely replicating the precise distribution of the instructions to generating answers that leverage the broader knowledge and capabilities of the pretrained base model.

An immediate and evident outcome of this strategy is that the model begins to produce responses that are not only more coherent but also lengthier. Both human evaluators and automated systems generally prefer longer, more detailed responses in most datasets, a phenomenon supported by the findings of \citet{dubois2023alpacafarm-rlhf}. However, our analysis suggests that the observed increase in verbosity only partially explains the improvements. The application of \symnoise{} results in enhancements in both the quality and quantity of responses, surpassing the achievements of \neftune{}. Notably, the average response length with \symnoise{} also exceeds that observed with the \neftune{} method. \symnoise{} can be conceptualized as assigning a non-zero probability to various noisy versions of the instruction set. Consequently, this strategy prompts the model to adopt a more generalized and uniform probability distribution, as opposed to a narrowly focused, peaky distribution that is overfitted to the training datasets—or a singular noisy variant of it. Such a methodology inherently fosters a more comprehensive understanding and application of the instruction set, thereby significantly reducing the risks associated with overfitting. 

\subsubsection{Longer responses vs repetition}\label{length_vs_rep}
In this section, our objective is to determine whether the lengthier responses produced using \symnoise{} are a result of increased repetition or if they contribute to more diverse and detailed content. 

Echoing the insights from \citet{jain2023neftune} and supporting evidence from leaderboard performances, a notable correlation emerges between extended response lengths and improved performance in the \texttt{AlpacaEval} task. This raises the question of whether the augmentation of response length by \symnoise{} could lead to diminished text diversity and quality. Our analysis scrutinized the frequency of N-gram repetitions in responses generated by \llama{}-2, trained on various datasets, both with and without \symnoise{} application.

Following the methodology of \citet{jain2023neftune}, our analysis was restricted to the initial segments of each response to maintain consistency. Specifically, we examined the first $50$ words for Alpaca-trained models, $100$ words for Evol-Instruct, $150$ words for ShareGPT, and $150$ words for OpenPlatypus, ensuring that at least half of the responses exceeded these thresholds. Responses shorter than these limits were excluded from the analysis.

To assess text diversity, we report \emph{log-diversity} in Table \ref{tab:length_vs_rep}, as described in~\cite{kirchenbauer2023reliability-logdiversity1, li2022contrastive-logdiversity2}. This metric evaluates the repetition rates of 2-, 3-, and 4-grams, acts as a gauge for text diversity and quality.

As delineated in Table \ref{tab:length_vs_rep}, the findings reveal that \symnoise{} typically yields lengthier responses. However, importantly, the frequency of 2-gram repetitions and overall token log-diversity remain largely consistent, paralleling the results observed with \neftune{}. This suggests that the increased length of responses under \symnoise{} is not simply due to repetitive content, but rather indicates the inclusion of additional, relevant information, thereby enriching the depth and value of the generated responses.

\subsubsection{Ablation study with different strength of noise}\label{noise_ablation}
In our final analytical step, we explored the efficacy of employing different noise distributions, specifically uniform~(\neftune{}) versus Gaussian noise, versus within the \symnoise{}{} algorithm. Our findings, detailed in Table~\ref{tab:noise_ablation}, reveal that Gaussian noise tends to produce longer outputs. However, this increased length does not correlate with a corresponding enhancement in performance. While it is generally observed that longer generations are associated with improved scoring, none of the generation-time strategies employed matched the effectiveness of models trained with \symnoise{}. Interestingly, our innovative approach, \symnoise{}, exhibits superior performance, surpassing benchmark results. It demonstrates an approximate improvement of $6.7\%$ over the models utilizing \neftune{}. Furthermore, we conducted a comparative analysis with Bernoulli noise to underscore the effectiveness of the symmetric opposing noise component in \symnoise{}.

\begin{table*}[h]
\centering
\caption{\texttt{AlpacaEval} Win Rate (alongside Average Character Count) assessed by ChatGPT across various strengths and types of training noise. Although character length tends to rise with increased noise, it does not consistently correlate with the Win Rate in \texttt{AlpacaEval}. We report base model and \neftune{} performance from \citet{jain2023neftune}. We notice that the avg. character length is slightly more than \neftune{} but less than Gaussian noise, yet performance is better than both the methods. We ran it with default parameters ($noise=5$) due to computation constraints, but still get best performance than all other parameter tuning.}\label{tab:noise_ablation}

\begin{tabular}{|l|c|c|c|c|c|c|} \toprule
Setting   & \multicolumn{2}{c}{Alpaca} &  \multicolumn{2}{c}{Evol-Instruct}   & \multicolumn{2}{c}{OpenPlatypus}      \\ \midrule
\llama{}-2-7b        & 48.26 & (375.22) & 62.55& (864.06)  & 57.20 & (1100.98) \\ \midrule
+\neft{} Noise 5  & 62.55& (1061.89)   & 67.58 & (1403.59) & 60.99& (1428.31) \\\midrule
+\neft{} Noise 10 & 61.18 & (1009.94)    & 65.59 & (1696.88) & 60.62 & (1833.85)  \\\midrule
+\neft{} Noise 15 & 61.86 & (819.61)    & 66.58 & (1650.65) & 61.74 & (1694.26) \\\midrule
+Gaussian Noise 5 & 60.93 & (1371.32)    & 65.09 & (2065.75) & 59.13 & (2060.92) \\  \midrule
+Bernoulli Noise 5 & 61.32 & (1272)    & 65.10 & (1840) & 60.22 & (1968) \\ \midrule
+\symnoise{} Noise 5 & 64.92 & (1186)    & 69.62 & (1700) & 62.14 & (1689) \\ \bottomrule
\end{tabular}
\end{table*}

\begin{table*}[bt!]
\centering
\caption{Following~\cite{jain2023neftune}, we report Average character counts~(Row 1) in \texttt{AlpacaEval} responses generated by \llama{}-2 models, each fine-tuned on distinct datasets. Alongside, we present the mean output length for these datasets, despite the training constraint of a maximum sequence length of 512 as suggested in~\cite{jain2023neftune}. Notably, \neft{} and \symnoise{} both enhance the average length, with \symnoise{} having slightly more length. (Row 2) Lengths determined by counting tokens separated by whitespace in generated text. (Row 3) Rates of repetition for 2-gram sequences. Results for \llama{}-2 and \neft{} are reported from~\citet{jain2023neftune}.}
\label{tab:length_vs_rep}
\begin{tabular}{|l|l|c|c|c|c|}
\toprule
&  & \shortstack{Alpaca\\($\alpha=5$)}      & \shortstack{Evol-Instruct\\($\alpha=5$)} & \shortstack{ShareGPT\\($\alpha=10$)}  & \shortstack{OpenPlatypus\\($\alpha=15$)} \\ \midrule
\multirow{3}{*}{\shortstack{\textbf{Character}\\\textbf{Lengths}}} & Training data     & 270.31     & 1356.43            & 1276.76   & 649.39 \\\cmidrule{2-6}
& \llama{}-2 7B     & 375.22     & 864.06            & 1011.28   & 1100.98 \\ 
& +\neft{} & 1061.89    & 1403.59           & 1496.86   & 1694.26 \\ 
& +\symnoise{} & 1186    & 1700           & 1584.72   & 1689 \\\midrule

\multirow{2}{*}{\shortstack{\textbf{Whitespace}\\\textbf{Lengths}}} & \llama{}-2 7B     & 60.5       & 138.99            & 161.04    & 170.41 \\ 
& +\neft{} & 169.36     & 225.56            & 234.99    & 264.12 \\
& +\symnoise{} & 176.26    & 245.50           & 255.86   & 270.26 \\ \midrule
\multirow{2}{*}{\shortstack{\textbf{2-Gram}\\\textbf{Repetition $\%$}}} & \llama{}-2 7B    & 1.49 & 3.87 & 4.82 & 2.73 \\ 
& +\neft{} & 1.72 & 3.79 & 4.58 & 3.21 \\ 
& +\symnoise{} & 1.80    & 3.80           & 4.86   & 3.30 \\ \bottomrule  
\end{tabular}
\end{table*}

\section{CONCLUSIONS AND LIMITATIONS}
The advancements showcased by \symnoise{} in training large language models (LLMs) emphasize the importance of innovative algorithmic strategies and regularization techniques. Echoing the sentiments of \cite{jain2023neftune}, the field of LLMs, unlike its counterpart in computer vision, has often favored standardized training methods focusing on model scaling and dataset expansion. However, \symnoise{} underscores the potential of fine-tuning techniques in enhancing model performance, particularly in situations where overfitting to limited instruction datasets is a concern.

Our research contributes a novel perspective to LLM training by implementing symmetric noise in the embedding process. This technique not only combats overfitting but also significantly boosts the model's performance, as demonstrated by the remarkable improvement in \texttt{AlpacaEval} scores. While standard methods achieved a score of 29.79\% when fine-tuning the \llama{}-2-7B model using Alpaca, \symnoise{} elevated this to 69.04\%, a 6.7\% enhancement over \neftune{}.

In conclusion, \neftune{} has made significant strides in fine-tuning language models (LLMs) through regularization methods. However, its underlying mechanisms are not entirely intuitive or transparent. Advancing this line of research, we introduced \symnoise{}, an approach that is more intuitive and interpretable than \neftune{}, and also demonstrates superior performance across various benchmarks. \symnoise{} thus marks a considerable progression in the fine-tuning of LLMs, emphasizing the necessity to explore innovative training methodologies. Our research not only deepens the understanding of regularization techniques but also aids in evolving language models to be more versatile and proficient, setting a foundation for future explorations in this field.

Despite these promising results, our study does face limitations. The reliance on \texttt{AlpacaEval} as the primary evaluation metric and the constraints of computational resources limited our ability to fully explore \symnoise{}'s effectiveness across various model sizes and datasets. These factors underscore the need for further research to deepen our understanding of \symnoise{} and its potential applications in LLMs.
\bibliography{main}
\bibliographystyle{icml2021}
\end{document}